\documentclass{article}

\usepackage[preprint, nonatbib]{./nips_2018}

\usepackage[utf8]{inputenc} 
\usepackage[T1]{fontenc}    
\usepackage{hyperref}       
\usepackage{url}            
\usepackage{booktabs}       
\usepackage{amsfonts}       
\usepackage{amsmath}
\usepackage{./algorithm}
\usepackage{./algorithmicx}
\usepackage[noend]{./algpseudocode}

\DeclareMathOperator*{\argminA}{arg\,min}

\usepackage{./nicefrac}       
\usepackage{./lineno}       
\usepackage{graphicx}

\title{Controlling the privacy loss with the input feature maps of the layers in convolutional neural networks}

\author{
  Woohyung Chun \\
  System LSI Division \\
  Samsung Electronics \\
  \texttt{wh.chun@samsung.com} \\
  \And
  Sung-Min Hong \\
  System LSI Division \\
  Samsung Electronics \\
  \texttt{sm79.hong@samsung.com} \\
  \And
  Junho Huh \\
  System LSI Division \\
  Samsung Electronics \\
  \texttt{huhjunho@samsung.com} \\
  \And
  Inyup Kang \\
  System LSI Division \\
  Samsung Electronics \\
  \texttt{inyup.kang@samsung.com}
}

\begin{document}

\maketitle

\begin{abstract}
We propose the method to sanitize the privacy of the IFM(Input Feature Map)s that are fed into the layers of CNN(Convolutional Neural Network)s. 
The method introduces the degree of the sanitization that makes the application using a CNN be able to control the privacy loss represented 
as the ratio of the probabilistic accuracies for original IFM and sanitized IFM.
For the sanitization of an IFM, the sample-and-hold based approximation scheme is devised to satisfy an application-specific degree of the sanitization.
The scheme approximates an IFM by replacing all the samples in a window with the non-zero sample closest to the mean of the sampling window.
It also removes the dependency on CNN configuration by unfolding multi-dimensional IFM tensors into one-dimensional streams to be approximated.
\end{abstract}

\section{Introduction} \label{sec:Intro}

CNN(Convolutional Neural Network) has become the representative DNN(Deep Neural Network) since it overwhelms other techniques in the contest of image recognition~\cite{alex} and its recent advances~\cite{googlenet, resnet} outperform the human beings in the image classification.
However, it is not clearly explained of what makes CNN better than human beings' recognition. 
Thus, as presented in~\cite{sexual_orientation}, if CNN is used to leak a very private matter of individuals, it is very hard to prevent CNN from exposing the privacy because we do not know what noise can hamper the recognition of a CNN. Actually, some noise that does not affect humans' recognition does make a huge difference for CNN~\cite{intriguing}. Also, some representation that does not make any sense for human may work for CNN~\cite{DNN_fool}.
One thing obvious is that CNN learns all the things from the data fed into it.
This implies what CNN recognizes as a noise comes from the data set.
In this paper, we mainly focus on how to add the effective noise that disrupts CNN's recognition for the purpose of privacy protection.

Differential privacy~\cite{cynthia} is the mathematical definition that has been introduced to measure the privacy loss for the domains that handle massive user data such as data mining applications. 
Google~\cite{rappor} and Apple~\cite{apple} have adopted the differential privacy techniques into their crowdsourced applications with some data sketching technique to approximate users' private data. 
However, the approximation techniques are devised to hide the privacy from the recognition of human beings. 
It would not be applicable to CNN unless the way that CNN learns from data is similar to that of human being.

The previous researches~\cite{dl_dp, semi_dp} focus on the control of privacy loss in the learning stage of DNN (particularly CNN since their experiments run on CNNs) rather than approximating the data that a DNN processes for inference.
Even if there is a way of hiding the privacy in the training stage, it would be difficult to force all the applications using CNNs to follow the way. 
Especially for the malicious attackers, the training method for the privacy-preservation can be the counter example of how CNN must not be trained for capturing the private sensitive information. 
Moreover, in the case that pre-trained CNNs are deployed on the personal devices such as mobile phone, the approach that controls the privacy loss in a training stage would not be applicable.

In order to protect the privacy from the potentially malicious CNNs that users cannot change (nor retrain in a privacy-preserving way), the privacy loss should be controlled in the level of FM(Feature Map) data that the layers of CNNs process. 
For controlling the privacy loss by manipulating the data that the layers of CNNs deal with, we need to resolve two difficulties: the various tensor dimensions according to network configurations and finding the control-knob that makes sure of lowering down the probabilistic accuracy.
To remove the dependency on various tensor dimensions, this research translates the multi-dimensional IFM(Input Feature Map) tensors into one-dimensional streams before they are approximated for the privacy preservation.

The most difficult part finds the control-knob that reduces the probabilistic accuracy monotonically in one direction 
(i.e. no accuracy increment happens when the accuracy is controlled to keep going down). 
However, due to the limitation of the training dataset, CNN cannot learn all the possible noises for a specific FM.
Thus, when FMs are approximated, the accuracy increase would be observed in a particular range even if the overall trend of accuracy goes down.
Instead of finding the best control-knob for a specific FM, this paper proposes the condition that any control-knob (that approximates FMs) should satisfy.
Thanks to the condition, even with the bad control-knob that keeps oscillating the probabilistic accuracy up and down, we can preserve the privacy of CNNs in a certain level that the condition designates.

This paper is organized as follows:
Section~\ref{sec:problem_description} describes the problem that controls the privacy loss of CNN with the IFMs of layers.
In Section~\ref{sec:proposed}, the degree of sanitization is introduced as the boundary condition that the method of decreasing the probabilistic accuracy should satisfy. Also, the IFM approximation scheme that reduces the accuracy and its network-wise control method are proposed.
Section~\ref{sec:evaluation} evaluates the proposed scheme on the layers of AlexNet in Caffe~\cite{caffe} CNN framework.
Finally, Section~\ref{sec:conclusion} concludes with the summary of our contribution.

\section{Problem Description} \label{sec:problem_description}

In the traditional signal processing, the noisy signal $\hat{s}(t)$ can be simply represented as the addition of the original signal $s(t)$ and noise $N(t)$. 
In case that $N(t)$ is clearly distinguished by a certain condition such as the passband frequency, $s(t)$ can be reconstructed from $\hat{s}(t)$ by filtering out $N(t)$. However, CNN learns the condition that distinguishes $s(t)$ and $\hat{s}(t)$ from the data fed in the training stage. Thus, it does not know all other noises which are not found in the data for training. 
To this end, it is not guaranteed that adding random noises to IFMs decreases the probabilistic accuracy of a CNN.

The ratio between the probabilistic accuracy of an original IFM and that of noisy IFM can be represented as the privacy loss, $\epsilon$ that is defined by the differential privacy~\cite{cynthia}. Intentionally adding noise to data for the purpose of hiding private information is called "sanitization" process. We primarily focus on the problem of sanitizing the IFMs for a CNN and it can be formulated by substituting CNN terms for the differential-privacy ones:
\begin{align}
	\frac{Pr[F(d) \in S]}{Pr[F(d')\in S]} \leq \exp(\epsilon) \label{eqn:DP_def}
\end{align}
where the randomized function $F$ is the last layer operation (e.g. softmax) of a CNN, $d$ is the input of the last layer of a CNN, $d'$ is the sanitized version of the input of the last layer in a CNN and $S$ is the subset of the label set $L$, whose elements have some probabilistic accuracies. The sanitization is only valid if and only if $\epsilon \geq 0$ in equation \ref{eqn:DP_def}.
In order to  make $\epsilon$ of equation  \ref{eqn:DP_def} $\geq$ 0, the set of $F(d')$ can be made as the subset of $F(d)$ 
\begin{align}
	\{F(d') | d' \in D \} \subseteq \{F(d) | d \in D\} \label{eqn:loss_con}
\end{align}
where $F: D \rightarrow L$. 
Equation \ref{eqn:loss_con} can be expressed in terms of the IFM for the layer before the last layer as below.
\begin{align}
	\{F(F'_{-1}(d_{-1}))|d_{-1} \in D_{-1} \} \subseteq \{F(F_{-1}(d_{-1})) | d_{-1} \in D_{-1} \} \label{eqn:approx_ofm}
\end{align}

where $F_{-1}$ is the layer before the last layer of a CNN, $d_{-1}$ is the IFM for the layer before the last layer and $F'_{-1}(d_{-1})$ means the sanitization of the OFM(Output Feature Map), $F_{-1}(d_{-1})$.
According to equation \ref{eqn:loss_con}, the left term of equation \ref{eqn:approx_ofm} can be replaced by 
$\{F(F_{-1}(d'_{-1}))|d'_{-1} \in D_{-1} \}$ because $F \circ F_{-1}$ can be regarded as a single function that has $d_{-1}$ as its IFM. That is,
\begin{align}
\{F(F_{-1}(d'_{-1}))|d'_{-1} \in D_{-1} \} \subseteq \{F(F_{-1}(d_{-1})) | d_{-1} \in D_{-1} \} \label{eqn:approx_con}
\end{align}
Equation \ref{eqn:approx_ofm} describes the relation between the sets of original OFM and sanitized OFM for $F_{-1}$. However, equation \ref{eqn:approx_con} shows the relation between the original IFM and sanitized IFM for $F_{-1}$. Thus, equation \ref{eqn:approx_con} is better to represent the condition that the IFM sanitization in the layer before the last layer of a CNN should satisfy.
Suppose that the IFM $d_{-k}$ of the k-th layer from the last layer of a CNN is sanitized as $d'_{-k}$. By letting the function, $G$ have $d_{-k}$ as its input and the output of the last layer of a CNN as its output, the relation between the sets of $G(d_{-k})$ and $G(d'_{-k})$ can be represented as
\[\{G(d'_{-k}) | d'_{-k} \in D_{-k} \} \subseteq \{G(d_{-k}) | d_{-k} \in D_{-k}\} \]
By replacing $G$ with $F \circ F_{-1} \circ F_{-2} ... \circ F_{-k}$
\begin{align}
	\{F(F_{-1}(F_{-2}(...F_{-k}(d'_{-k})...))) | d'_{-k} \in D_{-k} \} \subseteq \{F(F_{-1}(F_{-2}(...F_{-k}(d_{-k})...))) | d_{-k} \in D_{-k}\} \label{eqn:layer_san}
\end{align}
Equation \ref{eqn:layer_san} implies that IFM sanitization can make its result as the part of what an original IFM results in without any change of CNN layers. In order to meet the equation \ref*{eqn:layer_san}, $d'_{-k}$ can be made by sampling $d_{-k}$. The sampling scheme assumes that all the samples of an original IFM contribute to the probabilistic accuracy of an input image. The assumption is valid if the privacy loss, $\epsilon$ increases as the number of the samples selected in an IFM decreases. In the following section, we present the sample-and-hold approximation to control the degree of the privacy loss.

\section{Proposed Method} \label{sec:proposed}
This section mainly discusses the way of controlling the privacy loss by sanitizing the IFMs of the layers in a CNN.
The degree of privacy loss can be differently configured according to the application using a CNN.
Section \ref{sec:degree_of_sanitization} introduces the degree of sanitization to select the better sanitization knob to satisfy a given privacy loss.
Also, Section \ref{sec:sample-and-hold} devises the sample-and-hold approximation that sanitizes IFMs in fine-grained accuracy levels.
Finally, Section \ref{sec:layer-wise-sample} proposes the overall scheme where sample-and-hold approximation is controlled by the degree of sanitization.

\subsection{Degree of sanitization} \label{sec:degree_of_sanitization}
\begin{figure} [htb]
	\centering
	\includegraphics[width=0.8\linewidth]{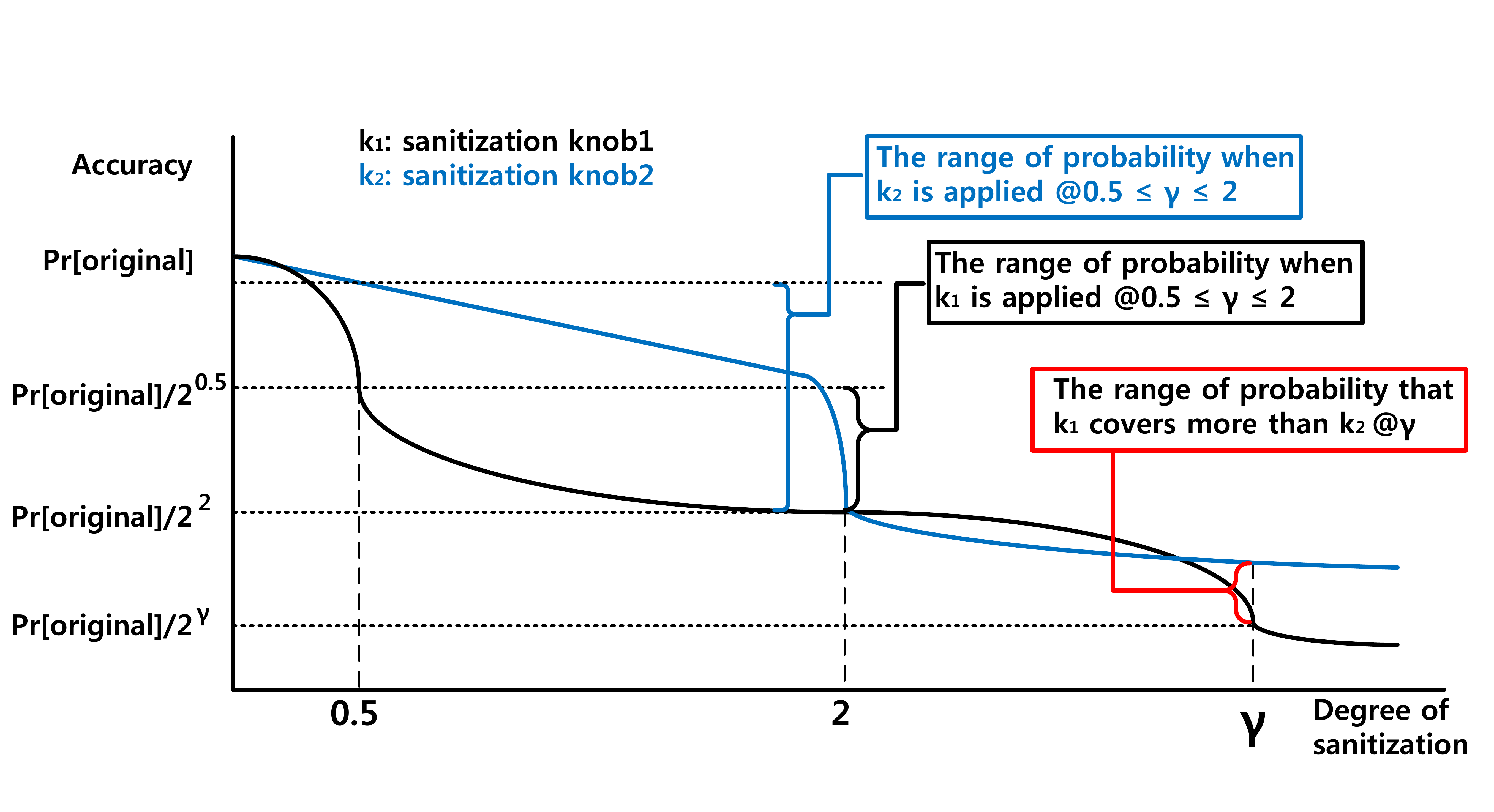}
	\caption{The degree of sanitization $\gamma$ determines the sanitization knob.}
	\label{fig:degree_of_sanitization}
\end{figure}

The application using a CNN needs to control the privacy loss $\epsilon$ of equation~\ref{eqn:DP_def}.
However, the term $exp(\epsilon)$ can have some loss when it is approximated as the rational number to work for the boundary condition.
Moreover, the loss is changing according to $\epsilon$.
In order to remove the loss from the boundary condition,
we introduce here the parameter called the degree of sanitization, $\gamma$ which linearly scales the strength of IFM sanitization in the layers of a CNN.
In equation \ref{eqn:DP_def}, the privacy loss $\epsilon$ can be translated into the linear equation having the constant slope, $ln2$ if the probability for the sanitized input is represented as $\frac{1}{2^{\gamma}}\times$ probability of the original input as shown in below:
\begin{align}
	\frac{Pr[F(d) \in S]}{Pr[F(d') \in S]} = \frac{Pr[F(d) \in S]}{{Pr[F(d) \in S]}/{2^{\gamma}} } &\leq \exp(\epsilon) \nonumber \\
	\gamma \times ln2 &\leq \epsilon 
	\label{eqn:degree_of_sanitization}
\end{align}
In equation \ref{eqn:degree_of_sanitization}, $\gamma$ determines the lower bound of the IFM sanitization. Figure \ref{fig:degree_of_sanitization} illustrates that $\gamma$ is used to evaluate the sanitization knob. In the figure, $Pr[\mbox{original}]$ corresponds to the probability of an original input, $Pr[F(d) \in S]$ in equation~\ref{eqn:degree_of_sanitization}.

In Figure \ref{fig:degree_of_sanitization}, when $\gamma$ ranges from 0 to 0.5, $k_{1}$ (sanitization knob1) is better than $k_{2}$ (sanitization knob2) since it is able to cover the full range from $\frac{Pr[\mbox{original}]}{2^{0.5}}$ to $Pr[\mbox{original}]$. 
In the same manner, in the $\gamma$ from 0.5 to 2, $k_{2}$ is better (i.e. it covers from $\frac{Pr[\mbox{original}]}{2^2}$ to somewhere beyond $\frac{Pr[\mbox{original}]}{2^{0.5}}$ and close to $P[\mbox{original}]$). 
However, for the given $\gamma$ that some application using a CNN might specify, $k_1$ is better than $k_2$ because it covers the probability range that $k_2$ does not include.
 
The sanitization method should decrease the probabilistic accuracy as the degree of sanitization increases. 
But, no CNN learns all the possible noises for a specific IFM due to the limitation of training dataset. So, there would be the cases of trend inversion where the probabilistic accuracy increases as the degree of sanitization increases. Then, we should pick the sanitization scheme that suppresses the trend inversion as much as possible. Section \ref{sec:sample-and-hold} develops the sanitization scheme that minimizes the trend inversion.

\subsection{Sample-and-hold approximation} \label{sec:sample-and-hold}
Each layer of a CNN deals with IFMs as the multi-dimensional tensors having a different size from other layers according to the way of stacking layers. 
In order to develop a sanitization scheme regardless of CNN structure and tensor dimension, IFMs need to be streamized before they are sanitized. 
That is, $n-$dimensional tensor $X_{{i_1},{i_2},...{i_n}} \in R^{I_1 \times I_2\times ... \times I_n}$ needs to be unfolded as the stream (i.e. one-dimensional tensor) $X_{i_1} \in {R_{s}}^{I_1}$ where $|{R_{s}}^{I_1}| = |R^{I_1 \times I_2 \times ... \times I_n}|$. 
We unfold the tensors in the direction that a layer function runs on an IFM (i.e. IFM width $\rightarrow$ IFM height $\rightarrow$ IFM channel).

\begin{figure} [htb]
	\centering
	\includegraphics[width=0.7\linewidth]{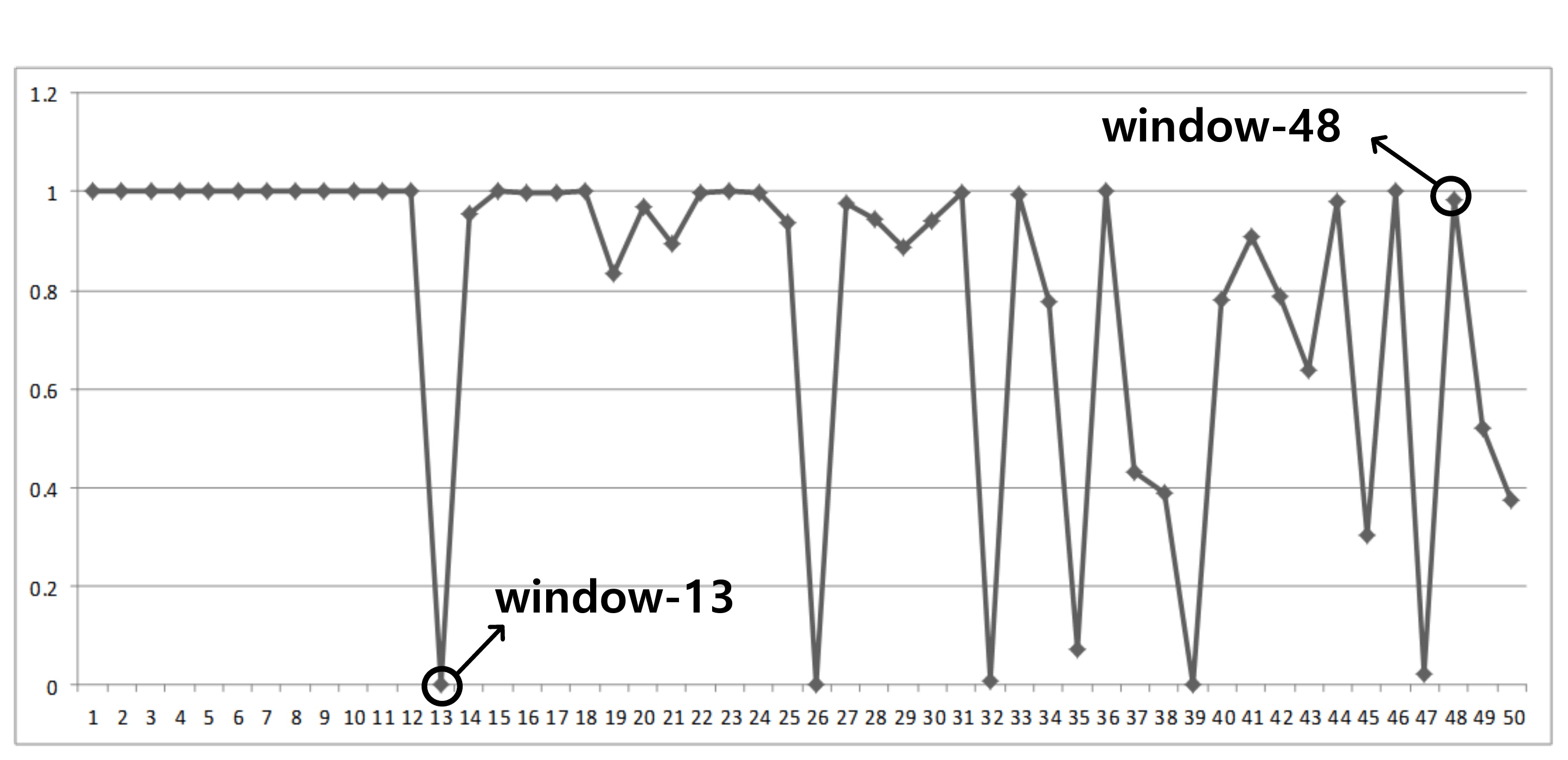}
	\caption{Wrong sampling does not follow the trend of the degree of sanitization.}
	\label{fig:wrong_sample_and_hold}
\end{figure}

Supposing that all the samples of an IFM contribute to the probabilistic accuracy, sample-and-hold method can decrease the probabilistic accuracy by reducing the number of distinct samples in an IFM (i.e. by increasing the size of a sampling window).
The size of a sampling window corresponds to the degree of sanitization $\gamma$ of sample-and-hold approximation.
To gradually decrease the probabilistic accuracy 
as the size of a sampling window increases, sample-and-hold method should decide which sample to select in a sampling window.
In case a wrong sample is selected in the window, the accuracy keeps oscillating even if the size of a sampling window increases (i.e. the degree of sanitization increases) as shown in Figure \ref{fig:wrong_sample_and_hold}.

Figure \ref{fig:wrong_sample_and_hold} shows the probabilistic accuracy when the first sample is selected in each window of the IFM for pool5 layer in AlexNet~\cite{alex} when input image is the picture of king penguin.
$X-$axis is the size of a sampling window and $y-$axis is the probabilistic accuracy. 
Even though window-13 (i.e. the case that window size is 13) has about 3.7 ($\approx$ 48/13) $\times$ more samples than window-48 (i.e. the case that window size is 48), the accuracy for window-13 is almost zero but that of window-48 is close to one. This implies that the samples of an IFM do not evenly contribute to the probabilistic accuracy. That is, the case of window-48 has some samples that the case of window-13 does not have and the samples contribute to the probabilistic accuracy much more than other samples. 

In order for the sample-and-hold approximation to reflect the importance of a sample, we develop the method that selects the sample which is the closest to the average among the non-zero samples within a window. The reason why average is computed only among non-zero samples is to prevent that the least sample is always selected in the layers that zeros are dominant. Algorithm~\ref{alg:around_mean_sample} summarizes the proposed sample-and-hold approximation.

\begin{algorithm}
	\caption{Proposed sample-and-hold approximation}
	\label{alg:around_mean_sample}
	\begin{footnotesize}
	\begin{algorithmic}[1]
		\State Let the window having n samples be $W = \{w_0, w_1, ..., w_{n-1}\}$
		\State Make $X = \{ x_p | x_p \in W_n, x_p \neq 0, 0 \leq p \leq (n-1) \} $
		\State if($X \neq \emptyset$)
		\State ~~~Find $E[X]$ // Calculate the average among the non-zero samples in a window
		\State ~~~// Find the sample closest to the non-zero average and let the sample represent an entire window
		\State ~~~$w_q =$ $\argminA_{x \in X} (|x - E[X]|)$, $w_q \in W$ 
		\State ~~~$W = \{w_{r}|w_{r} = w_{q}, r = 0,1, ..., n-1 \}$
		\State endif   
	\end{algorithmic}
	\end{footnotesize}
\end{algorithm}

Algorithm~\ref{alg:around_mean_sample} just describes the procedure applied to a single window. 
The overall sanitization scheme when the degree of sanitization $\gamma$ is given as a specification (i.e. constraint for a privacy loss) is discussed in in Section \ref{sec:layer-wise-sample}.

\subsection{Layer-wise sample-and-hold sanitization} \label{sec:layer-wise-sample}

\begin{figure} [htb]
	\centering
	\includegraphics[width=0.9\linewidth]{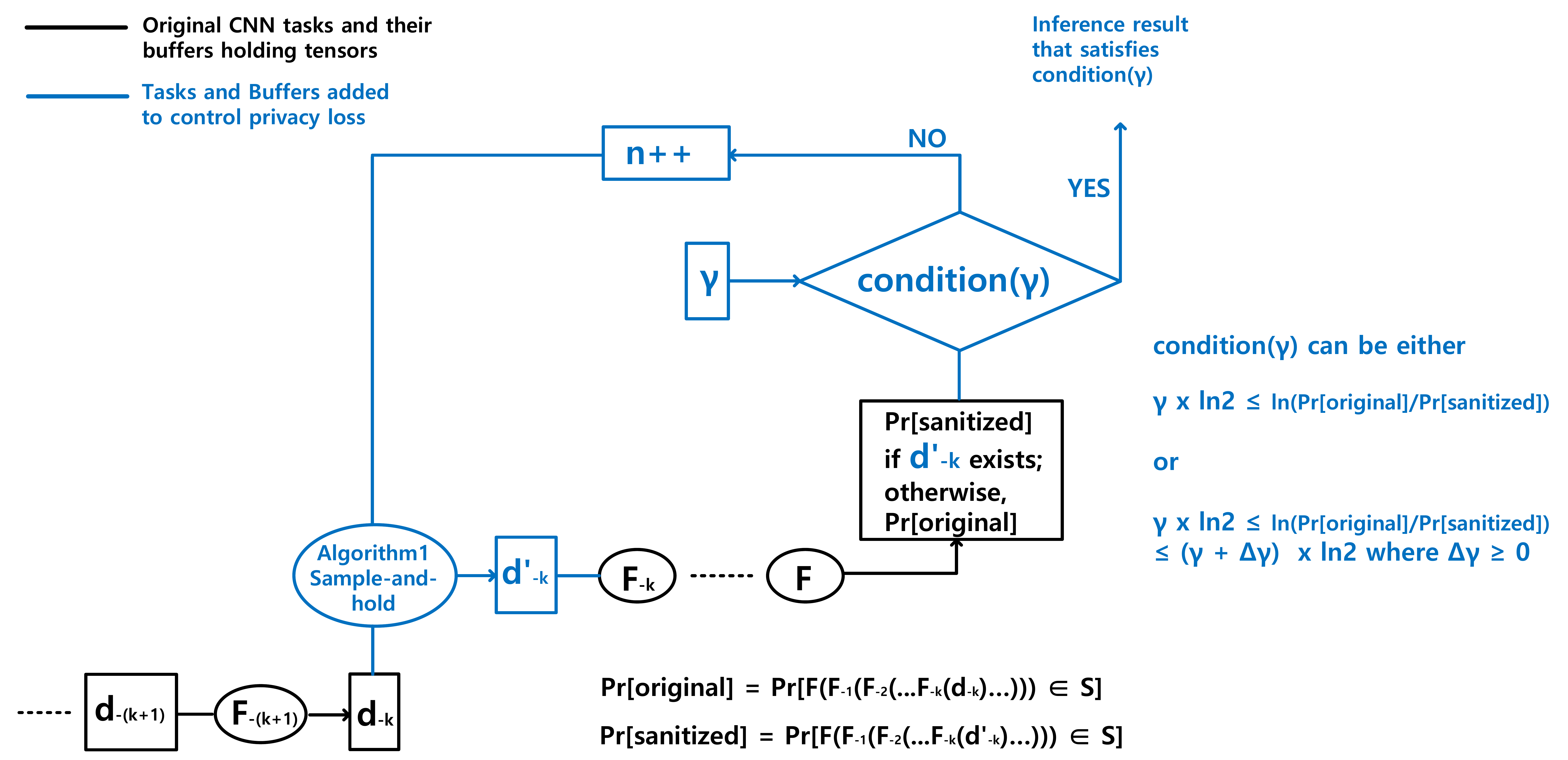}
	\caption{Layer-wise sanitization of an input feature map to meet the condition that a given degree of sanitization $\gamma$ describes}
	\label{fig:layerwise_sanitization}
\end{figure}

Figure~\ref{fig:layerwise_sanitization} shows how Algorithm~\ref{alg:around_mean_sample} is applied for sanitizing 
the IFM of a CNN layer to satisfy the condition that 
a given degree of sanitization $\gamma$ specifies. 
The blue colored tasks and buffers are required to be added for the layer-wise sanitization using Algorithm~\ref{alg:around_mean_sample}. 
In the figure, $F$ is the last layer of a CNN, $n$ is the number of samples in a window and $\gamma$ is the degree of sanitization.
Also, $F_{-k}$ is the k-th layer frmo the last layer of a CNN, $d_{-k}$ is the IFM of the k-th layer and $d'_{-k}$ is the sanitized version of $d_{-k}$.
The buffer for
$\gamma$ and the task of checking if the inference result meets the degree of sanitization
work for an entire CNN.
However, the sample-and-hold task and its output buffer having $d'_{-k}$, and the counting buffer "$n++$" which increases the size of a sampling window are required for every layer in a CNN.

In Figure~\ref{fig:layerwise_sanitization}, the blue parts should be realized on the platform-level inference software (e.g. Caffe inference) in order to invalidate any malicious attempt by changing the network configuration (e.g. skipping the sanitization by replacing $d'_{-k}$ with $d_{-k}$).
The implementation is feasible since our proposed scheme deletes the network dependency by
translating the multi-dimensional IFM tensors into one-dimensional streams. 
However, before the sanitized streams are fed into the next layer, they should be tensorized to have the same dimension with the original IFM.
To this end, the operational dependency among the edges of the task "Algorithm~\ref{alg:around_mean_sample}" must be maintained as below:
\begin{align}
start\_read_{n++} < start\_read_{d_{-k}} < start\_write_{d'_{-k}} 
\label{eqn:op_dep}
\end{align}
where $start\_read_{n++}$ is the start time of reading data from the counting buffer marked as "$n++$" in Figure~\ref{fig:layerwise_sanitization}, $start\_read_{d_{-k}}$ is the start time of reading data from the buffer having the original IFM $d_{-k}$ and $start\_read_{d'_{-k}}$ is the start time of writing data to the buffer for the sanitized IFM $d'_{-k}$.
Proposed sample-and-hold sanitization provides the different distribution of probabilistic accuracies according to layers because the approximation selecting the sample closest to the average among the non-zero samples is affected by the ratio of zeros and the sparsity of non-zero samples.
In the next section, we explore the layer-wise aspects of the proposed method through some evaluation metric.

\section{Evaluation of Proposed Method} \label{sec:evaluation}

\begin{figure} [htb]
	\centering
	\includegraphics[width=0.7\linewidth]{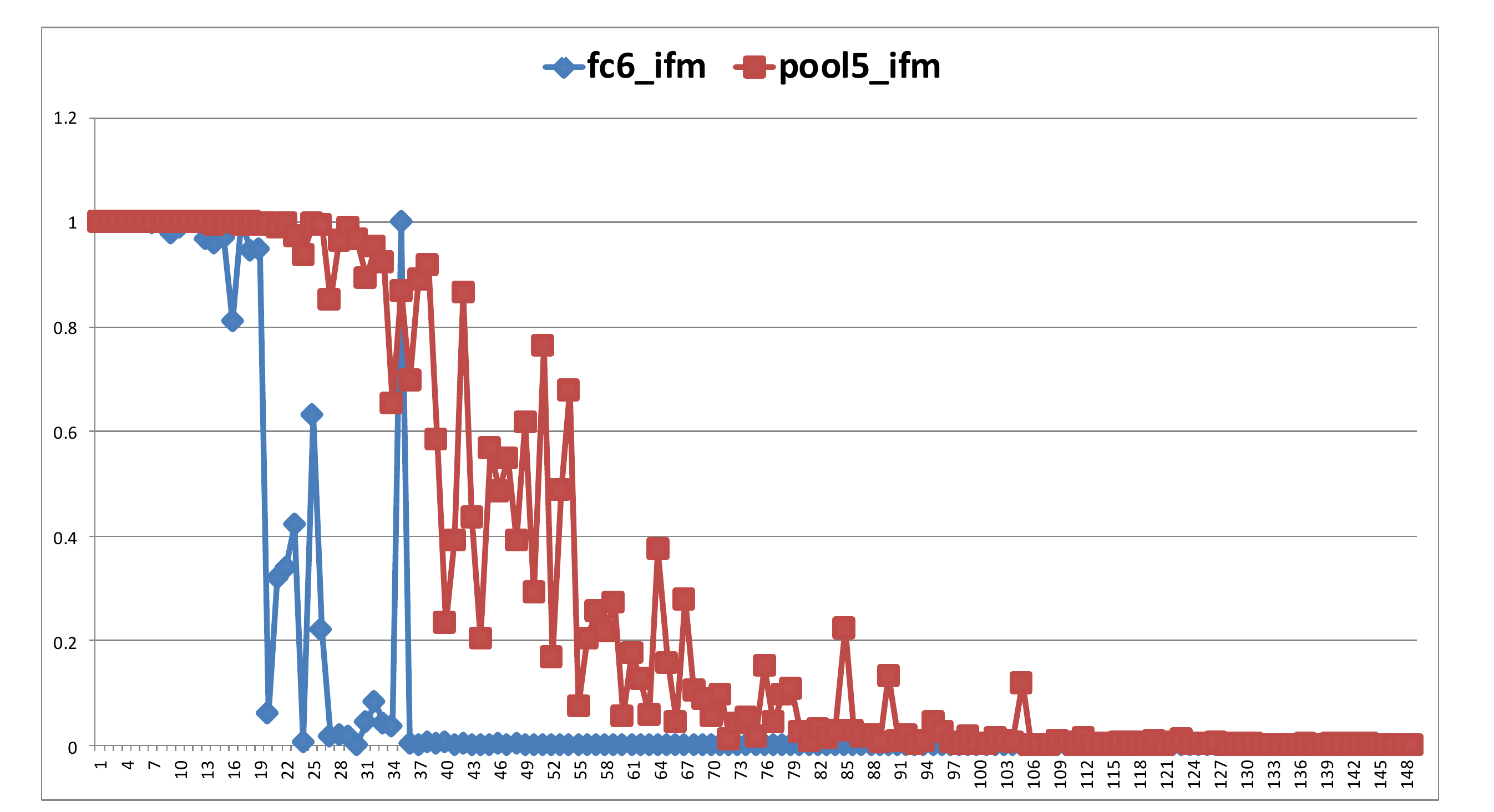}
	\caption{Sanitization of pool5 and fc6 input feature maps in AlexNet}
	\label{fig:efficient_approximation}
\end{figure}

We evaluate the proposed sample-and-hold approximation in the layers of AlexNet~\cite{alex}.
Figure~\ref{fig:efficient_approximation} shows two different IFMs are sanitized respectively by the proposed sample-and-hold approximation when a picture of king penguin is fed into AlexNet for inference. $X-$axis indicates the size of sampling window and $y-$axis notes the probabilistic accuracy. Larger sampling window degrades probability accuracy more.

In Figure~\ref{fig:efficient_approximation}, the sanitization of pool5 IFM provides more steps for the probability reduction than the case of sanitizing fc6 IFM does. 
In order to quantify the efficiency of the sanitization, we need to measure the following ratio:
\begin{align}
\mbox{eff}_{\mbox{san}} = \frac{\mbox{The number of different probabilistic accuracies}}{\mbox{The range of of sampling-window sizes}}
\label{eqn:eff_metric}
\end{align}
According to equation~\ref{eqn:eff_metric}, the sanitization of pool5 IFM is more efficient than that of fc6 IFM because $\mbox{eff}_{\mbox{san}}\mbox{(fc6\_ifm)} = \frac{42}{149} \approx 0.28$ and $\mbox{eff}_{\mbox{san}}\mbox{(pool5\_ifm)} = \frac{120}{149} \approx 0.81$.
Table~\ref{table:alex_san} lists $\mbox{eff}_{\mbox{san}}$s for all the IFMs of AlexNet when the proposed sample-and-hold approximation which scales its window size from 2 to 150 is applied to the case where the picture of a king penguin is fed as the input image.
It also breaks down the range of probabilistic accuracies.
Our sample-and-hold approximation selects the non-zero sample having the minimum distance from the mean value among non-zero samples and it replaces the sample with all others in a window. 
Thus, in an IFM, if the ratio of zeros is high and non-zero samples are densely populated, $\mbox{eff}_{\mbox{san}}$ becomes high. 
Also, we can get a high $\mbox{eff}_{\mbox{san}}$ in the IFM having a lot of samples.

\begin{table}
	\caption{$\mbox{eff}_{\mbox{san}}$s for the input feature maps of AlexNet and the distribution of probabilistic accuracies}
		\centering
		\begin{scriptsize}
			\begin{tabular}{|c|l|l|l|l|l|}
				\hline
                              &                           & number of the different      &  number of the different      & number of the different        & Ratio of zeros      \\
	IFMs                  & $\mbox{eff}_{\mbox{san}}$ & probabilistic accuracies     &  probabilistic accuracies     & probabilistic accuracies       & in an original IFM  \\
				&                           & between 0.0 and 0.2          &  between 0.2 and 0.8        & between 0.8 and 1.0            & (= (number of zeros) /  \\
				&		            &				   &				 &				  &    (total number of samples) \\
				&			    &				   &				 &				  &    in an orignal IFM ) \\
				\hline
				conv1  & 0.07 	 			& 9 	& 0 	& 2 	& 0.0 								\\
               & ($\approx$ 11/149)  &     &     &     &	( $=$ 0/154587) 				\\
		    \hline
		    norm1  & 0.61 				& 88 	& 2 	& 1 	& 0.49 				\\
							 & ($\approx$ 91/149) 	&			&			&			&	($\approx$ 143153/290400) 	\\
		    \hline
		    pool1  & 0.22 				& 30 	& 1 	& 2 	& 0.49 				\\
									& ($\approx$ 33/149) 	&			& 1 	&			& ($\approx$ 143153/290400) 	\\
		    \hline
		    conv2  & 0.19 				& 27 	& 1 	& 1 	& 0.11  					\\
									& ($\approx$ 29/149)	&			&			&			& ($\approx$ 7912/69984)			\\
		    \hline
		    norm2  & 0.93 			& 100 & 28 	& 10 	& 0.78  				\\
									&	($\approx$ 138/149)	&			&			&			&	($\approx$ 146290/186624)		\\
				\hline
		    pool2  & 0.87 			& 115	& 4 	& 10 	& 0.78  				\\
									& ($\approx$ 129/149)	&			&			&			& ($\approx$ 146290/186624)		\\
				\hline
		    conv3  & 0.32 	 			& 37 	& 4 	& 6 	& 0.48 					\\
									& ($\approx$ 47/149)	&			&			&			& ($\approx$ 20874/43264)			\\
				\hline
				conv4  & 0.15 				& 18 	& 0 	& 4 	& 0.73 					\\
									& ($\approx$ 22/149)  &			&			&			& ($\approx$ 47512/64896)			\\
				\hline
				conv5  & 0.24 				& 29 	& 2 	& 5 	& 0.69 					\\
									&	($\approx$ 36/149)	&			&			&			& ($\approx$ 44785/64896)			\\
				\hline
				pool5  & 0.81 			& 70  & 23  & 27 	& 0.88 					\\
									&	($\approx$ 120/149)	&			&			&			&	($\approx$ 38128/43264)			\\
				\hline
				fc6  	& 0.28 			& 25  & 5  	& 12  & 0.62 					\\
									& ($\approx$ 42/149)	&			&			&			&	($\approx$ 5693/9216)				\\
				\hline
				fc7  	& 0.02 			& 2 	& 1 	& 0 	& 0.85 						\\
									&	($\approx$ 3/149)		&			&			&			& ($\approx$ 3469/4096)				\\
				\hline
				fc8  	& 0.04 			& 4 	& 0 	& 2 	& 0.81 					\\
									& ($\approx$ 6/149)		&			&			&			& ($\approx$ 3330/4096)				\\
				\hline
			\end{tabular}
		\end{scriptsize}
	\label{table:alex_san}
\end{table}

In Table~\ref{table:alex_san}, norm1 IFM and norm2 IFM tend to have densely populated non-zero samples and large amount of the zeros since both come from the consecutive convolution-relu operations.
pool1 IFM and pool2 IFM 
have the same zero-ratios with their corresponding norm IFMs (i.e. norm1 IFM for pool1 IFM and norm2 IFM for pool2 IFM). However, their operations reduce the number of zeros by replacing zeros with non-zero samples. Thus, their $\mbox{eff}_{\mbox{san}}$ becomes lower than their precedent norm IFMs'.
On the other hand, pool5 IFM which comes after successive convolution-relu pairs (i.e. conv3-relu3, conv4-relu4 and conv5-relu5) gets high $\mbox{eff}_{\mbox{san}}$ due to the densely populated non-zero samples and the high ratio of zeros.

Compared to other IFMs, both fc7 and fc8 IFMs have the small number of samples, 4096. 
This means the number of sampling windows cannot be more than 2048 (since the minimum size of sampling window is 2).  
For fc8 IFM, only length-2 and length-3 sampling windows go beyond the probabilistic accuracy of 0.8.
If the number of the different probabilistic accuracies $\geq 0.8$ for fc8 IFM is scaled up to the case that has the same number of samples with pool2 IFM,
(number of the different probabilistic accuracies $\geq$ 0.8) : (total number of samples in an IFM) = 2 : 4096 = 10 : $x$
and $x = 20480$ (< 186624 for pool2 IFM). This means that fc8 IFM can have larger number of the different probabilistic accuracies $\geq 0.8$ than pool2 IFM if it has the same number of samples with pool2 IFM.

\begin{figure} [htb]
	\centering
	\includegraphics[width=0.8\linewidth]{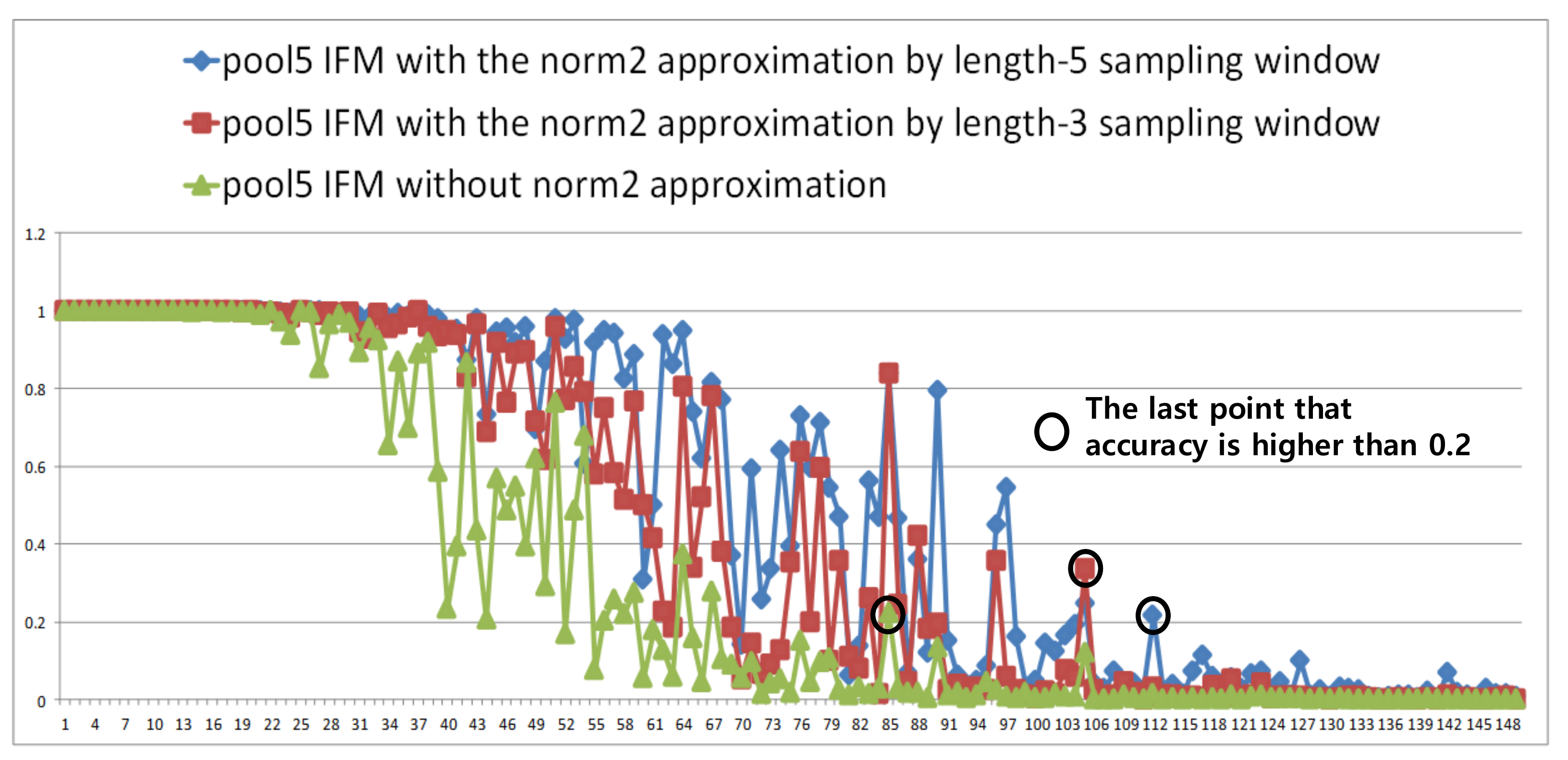}
	\caption{Larger sampling window makes a longer attenuation range.}
	\label{fig:multiIFM_sanitization}
\end{figure}

$\mbox{eff}_{\mbox{san}}$ can be enhanced by approximating the multiple IFMs.
If the approximation with a large sampling window does not change the accuracy of an original IFM,
the ineffectual non-zero samples (that come from the approximation) can give more granules of the probabilistic accuracy to the approximation of upcoming layers
because the proposed sample-and-hold scheme works only on non-zero samples. 
For example, in Table~\ref{table:alex_san}, norm2 IFM holds its probability as 1.0 until its sampling window becomes 5. 
When pool5 IFM is sanitized with the norm2 IFM approximated by length-3 ( or length-5 ) sampling window,
$\mbox{eff}_{\mbox{san}}$ and the distributions of probabilistic accuracies are changed as Table~\ref{table:multilayer_san}.
The norm2-IFM approximation enhances $\mbox{eff}_{\mbox{san}}$ of pool5 IFM and 
the approximation also tends to make pool5 IFM have more granules in the range of high probabilistic accuracies ($\geq 0.8$).
The more granules prolong the attenuation range of the probabilistic accuracies as shown in Figure~\ref{fig:multiIFM_sanitization}. 

\begin{table}
	\caption{$\mbox{eff}_{\mbox{san}}$s and the distribution of probabilistic accuracies when the proposed scheme sanitizes the pool5 input feature map after norm2 input feature map is approximated to have the same accuracy with the original input feature map}
		\centering
		\begin{scriptsize}
			\begin{tabular}{|l|c|c|c|c|c|}
				\hline
			   IFMs                       &                           & number of the different      &  number of the different      & number of the different          \\
					                          & $\mbox{eff}_{\mbox{san}}$ & probabilistic accuracies     &  probabilistic accuracies     & probabilistic accuracies         \\
			                               &                           & between 0.0 and 0.2          &  between 0.2 and 0.8          & between 0.8 and 1.0              \\
				\hline
				pool5 IFM after						&								 &		 &		 &		 \\
				original norm2 IFM				& 0.81 (120/149) & 70  & 23  & 27  \\
				         									&                &     &     &     \\
				\hline
				pool5 IFM	after  						&									&			&			&			\\
				norm2 IFM										&									&			&			&			\\
				approximated with						& 0.83 (125/149)	&	63	& 26	& 36	\\
				the length-3 window					&				    	    &     &     &     \\                        
				\hline
				pool5 IFM	after 						&									&			&			&			\\
				norm2 IFM										&									&			&			&			\\
				approximated with						& 0.90 (134/149)	& 57	& 28	& 49	\\
				the length-5 window					&				    	    &     &     &     \\                        
				\hline
			\end{tabular}
		\end{scriptsize}
	\label{table:multilayer_san}
\end{table}

In Figure~\ref{fig:multiIFM_sanitization}, $X-$axis is the size of sampling window and $y-$axis is the probabilistic accuracy. The probabilistic accuracy of "pool5 IFM without norm2 approximation" becomes less than 0.2 when the window size becomes larger than 86.
However, "pool5 IFM with the norm2 approximation by length-3 sampling window" goes below 0.2 for the window size $\geq 106$ and
"pool5 IFM with the norm2 approximation by length-5 sampling window" needs to have the sampling window longer than 113 to get its probabilistic accuracy lower than 0.2.

\section{Conclusion} \label{sec:conclusion}
In this paper, we proposed the sample-and-hold approximation scheme that sanitizes the privacy of the IFM(Input Feature Map)s that go through the layers of CNN(Convolutional Neural Network)s.
In order to remove the dependency on the network configuration coming from the various tensor dimensions, the proposed approximation unfolds the multi-dimensional IFM tensors into the one-dimensional stream.
And then, the scheme selects the non-zero sample having the minimum distance from the mean among the non-zero samples in a window, as the representative of the window to reflect the importance of a sample by the probability mass and value.

Also, we introduce the degree of the sanitization which works as the systematic boundary condition that prevents a certain amount of privacy from being leaked even in the case the proposed sample-and-hold approximation does not work well.
The proposed scheme is evaluated in the layers of AlexNet by the metric, the efficiency of the sanitization which is affected by the ratio of zeros, the density of non-zero samples and the number of samples in an IFM.

\end{document}